\documentclass{article}
\usepackage{spconf,amsmath,epsfig}
\usepackage{times}
\usepackage{color}
\usepackage{algorithmicx}
\usepackage[ruled]{algorithm}
\usepackage{algpseudocode}
\usepackage{url}
\usepackage{amssymb}
\usepackage{bbm}
\usepackage{subfigure}
\usepackage{graphicx} 
\usepackage{hyperref}

\graphicspath{{./}{./Figs/}}

\begin{document}\sloppy

\def\x{{\mathbf x}}
\def\L{{\cal L}}

\title{Robust and Real-time Deep Tracking Via Multi-Scale Domain Adaptation}
%
\name{Xinyu Wang$^1$, Hanxi Li$^{1\ast}$, Yi Li$^2$, Fumin Shen$^3$, Fatih Porikli$^4$}
\address{Jiangxi Normal University, China$^1$\\
        Toyota Research Institute of North America, USA$^2$\\
        University of Electronic Science and Technology of China, China$^3$\\
        Australian National University, Australia$^4$\\}
\maketitle

\begin{abstract}
Visual tracking is a fundamental problem in computer vision. Recently, some
deep-learning-based tracking algorithms have been achieving record-breaking performances.
However, due to the high complexity of deep learning, most deep trackers suffer from low
tracking speed, and thus are impractical in many real-world applications. Some new
deep trackers with smaller network structure achieve high efficiency while at the cost of
significant decrease on precision. In this paper, we propose to transfer the feature for
image classification to the visual tracking domain via convolutional channel reductions.
The channel reduction could be simply viewed as an additional convolutional layer with the
specific task. It not only extracts useful information for object tracking but also
significantly increases the tracking speed. To better accommodate the useful feature of
the target in different scales, the adaptation filters are designed with different sizes.
The yielded visual tracker is real-time and also illustrates the state-of-the-art
accuracies in the experiment involving two well-adopted benchmarks with more than 100 test
videos.
\end{abstract}
\section{Introduction}
\label{sec:intro}

Visual tracking is one of the long standing computer vision tasks. During the last decade,
as the surge of deep learning, more and more tracking algorithms benefit from deep neural
networks, e.g. Convolutional Neural Networks \cite{NIPS2013_5192,li2016deeptrack} and
Recurrent Neural Networks \cite{milan2016online,ning2016spatially}. Despite the
well-admitted success, a dilemma still existing in the community is that, deep learning
increases the tracking accuracy, while at the cost of high computational complexity. As a
result, most well-performing deep trackers usually suffer from low efficiency
\cite{ma2015hierarchical,nam2015learning}. Recently, some real-time deep trackers were
proposed \cite{held2016learning,bertinetto2016fully}. They achieved very fast tracking
speed, but can not beat the shallow methods in some important evaluations, as we
illustrate latter. 

\begin{figure}[thb]
\begin{centering}
\includegraphics[width=0.5\textwidth]{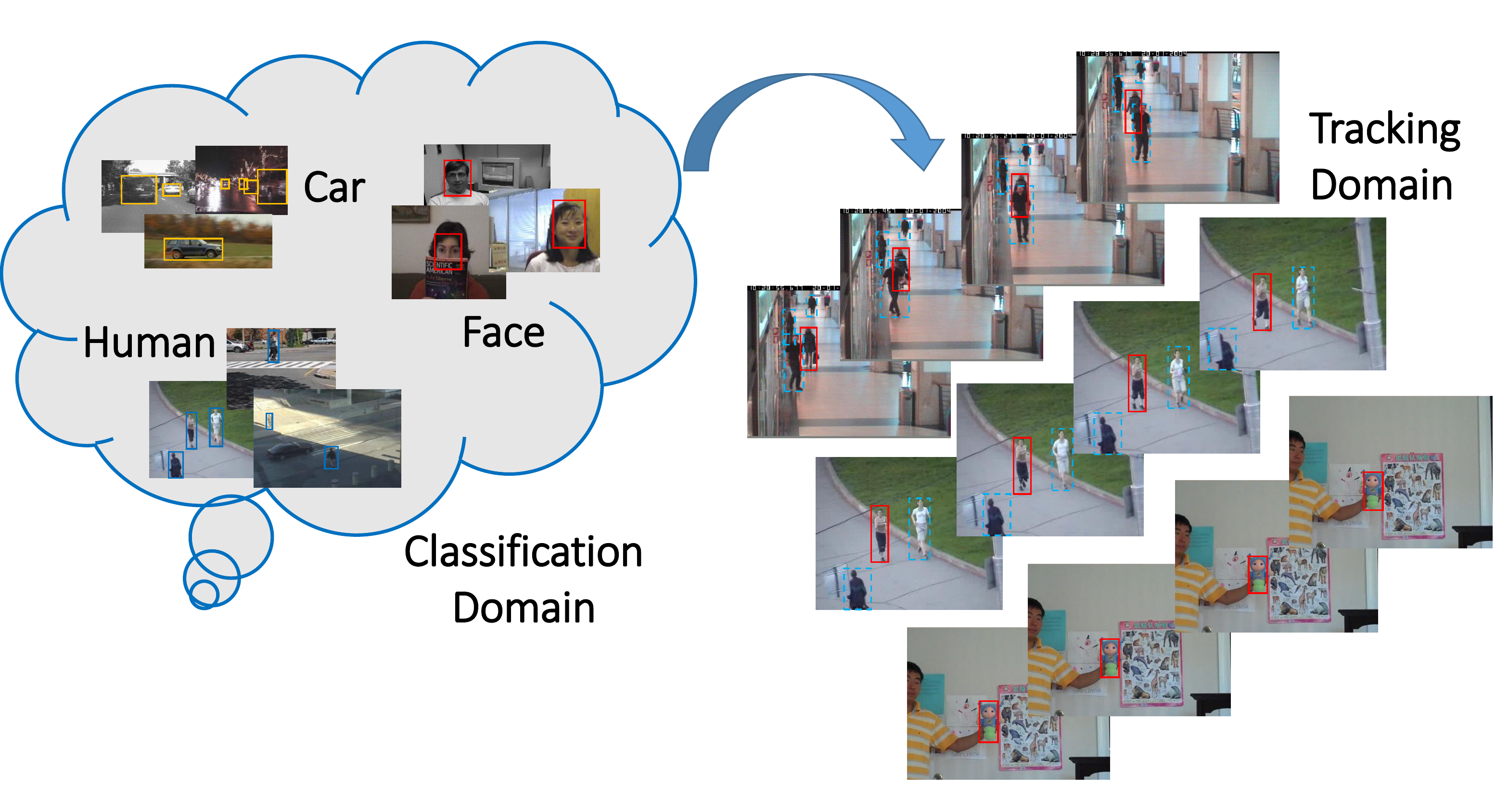}
\par\end{centering}
\caption{
  The high level concept of the proposed MSDAT tracker. Left: most of the deep neural network is
  pretrained for image classification, where the learning algorithm focus on object classes.
  Right: an adaptation is performed to transfer the classification features to the visual
  tracking domain, where the learning algorithm treats the individual object independently.
}  
\label{fig:adapdation}
\end{figure}

In this paper, a simple yet effective domain adaptation algorithm is proposed. The
facilitated tracking algorithm, termed Multi-Scale Domain Adaptation Tracker
(MSDAT), transfers the features from the classification domain to the tracking domain,
where the individual objects, rather than the image categories, play as the learning
subjects. In addition, the adaptation could be also viewed as a dimension-reduction
process that removes the redundant information for tracking, and more importantly, reduces
the channel number significantly. This leads to a considerable improvement on tracking
speed. Figure~\ref{fig:adapdation} illustrates the adaptation procedure.  To accommodate
the various features of the target object in different scales, we train filters with
different sizes, as proposed in the Inception network \cite{szegedy2015going} in the
domain adaptation layer. Our experiment shows that the proposed MSDAT algorithm runs in
around $35$ FPS while achieves very close tracking accuracy to the state-of-the-art
trackers. To our best knowledge, our MSDAT is the best-performing real-time visual
tracker.  

\section{Related work}
\label{sec:related}

Similar to other fields of computer vision, in recent years, more and more
state-of-the-art visual trackers are built on deep learning. \cite{NIPS2013_5192} is a
well-known pioneering work that learns deep features for visual tracking. The DeepTrack
method \cite{ourBMVC2014,li2016deeptrack} learns a deep model from scratch and updates it
online and achieves higher accuracy. \cite{wang2015transferring, icml2015_hong15} adopt
similar learning strategies, \emph{i.e.}, learning the deep model offline with a large
number of images while updating it online for the current video sequence.
\cite{zhang2015robust} achieves real-time speed via replacing the slow model update with a
fast inference process.

The HCF tracker \cite{ma2015hierarchical} extracts hierarchical convolutional features
from the VGG-19 network \cite{Simonyan14c}, then puts the features into correlation filters
to regress the respond map. It can be considered as a combination between deep learning
and the fast shallow tracker based on correlation filters. It achieves high tracking
accuracy while the speed is around $10$ fps. 
Hyeonseob Nam \emph{et al.} proposed to pre-train deep CNNs in multi domains, with each domain
corresponding to one training video sequence \cite{nam2015learning}. The authors claim  
that there exists some common properties that are desirable for target representations in
all domains such as illumination changes. To extract these common features, the authors
separate domain-independent information from domain-specific layers. The yielded
tracker, termed MD-net, achieves excellent tracking performance while the tracking speed
is only $1$ fps.

Recently, some real-time deep trackers have also been proposed. In
\cite{held2016learning}, David Held \emph{et al.} learn a deep regressor that can predict
the location of the current object based on its appearance in the last frame. The tracker
obtains a much faster tracking speed (over $100$ fps) comparing to conventional deep
trackers. Similarly, in \cite{bertinetto2016fully} a fully-convolutional siamese network
is learned to match the object template in the current frame. It also achieves real-time
speed. Even though these real-time deep trackers also illustrate high tracking accuracy,
there is still a clear performance gap between them and the state-of-the-art deep
trackers.

\section{The proposed method}
\label{sec:method}

In this section, we introduce the details of the proposed tracking algorithm, \emph{i.e.},
the Multi-Scale Domain Adaptation Tracker (MSDAT).

\subsection{Network structure}
\label{subsec:structure}

In HCF \cite{ma2015hierarchical}, deep features are firstly extracted from multiple layers
from the VGG-19 network \cite{Simonyan14c}, and a set of KCF \cite{henriques2015high}
trackers are carried out on those features, respectively. The final tracking prediction is
obtained in a weighted voting manner. Following the setting in \cite{ma2015hierarchical},
we also extract the deep features from $conv3\_5$, $conv4\_5$ and $conv5\_5$ network
layers of the VGG-19 model. However, the VGG-19 network is pre-trained
using the ILSVRC dataset \cite{ILSVRC15} for image classification, where the learning
algorithm usually focus on the object categories. This is different from visual tracking
tasks, where the individual objects are distinguished from other ones (even those 
from the same category) and the background. Intuitively, it is better to transfer the
classification features into the visual tracking domain. 

\begin{figure}[htb]
\begin{centering}
\includegraphics[width=0.45\textwidth]{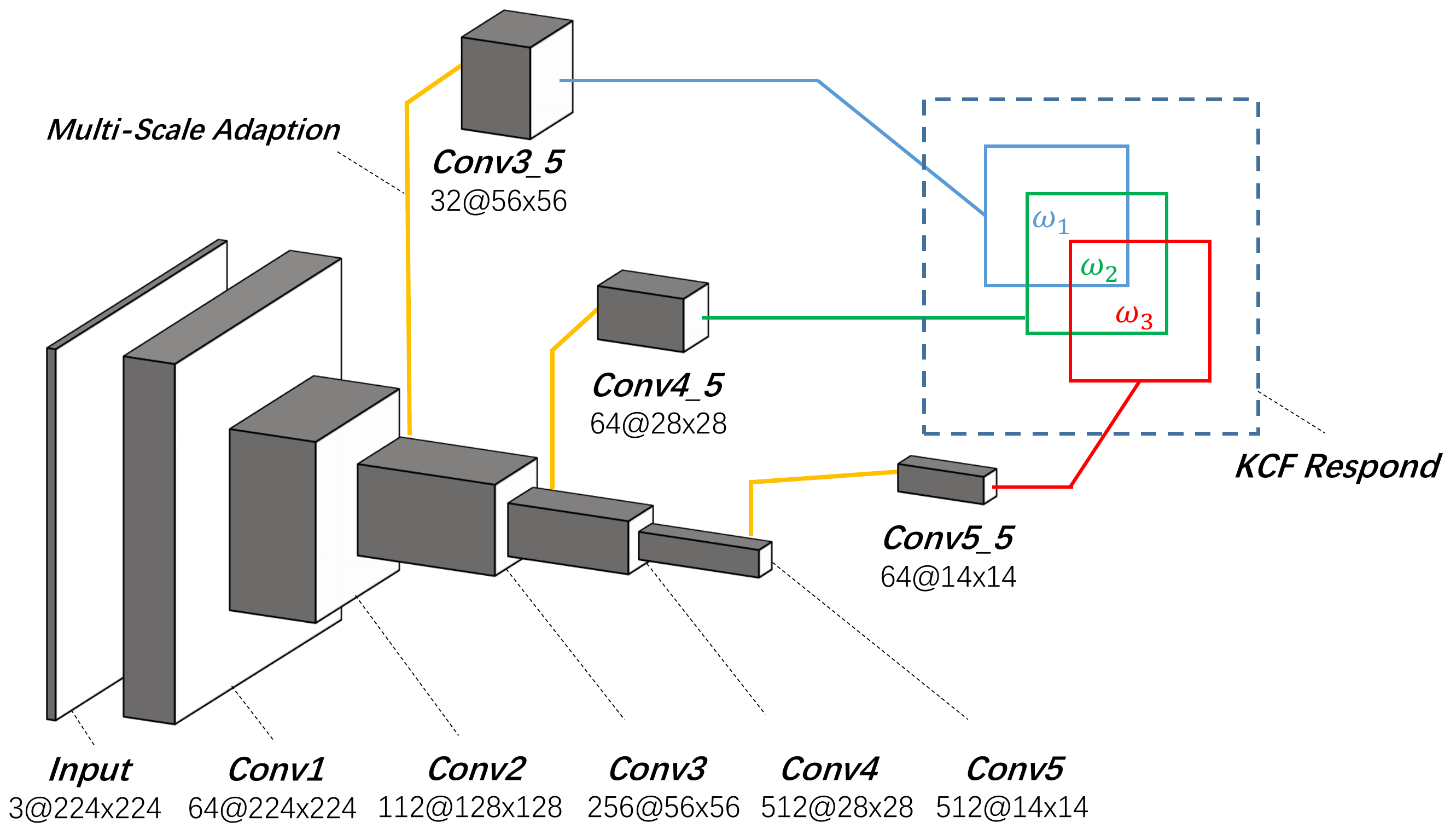}
\par\end{centering}
\caption{
  The network structure of the proposed MSDAT tracker. Three layers, namely, $conv3\_5$,
  $conv4\_5$ and $conv5\_5$ are selected as feature source. The domain adaption (as shown
  in yellow lines) reduces the channel number by $8$ times and keeps feature
  map size unchanged. Better viewed in color.
}  
\label{fig:structure}
\end{figure}
In this work, we propose to perform the domain adaptation in a simple way. A ``tracking
branch'' is ``grafted'' onto each feature layer, as shown in Fig.~\ref{fig:structure}. The
tracking branch is actually a convolution layer which reduces the channel number by $8$
times and keeps feature map size unchanged. The convolution layer is then learned via
minimizing the loss function tailored for tracking, as introduced below.

\subsection{Learning strategy}
\label{subsec:learn}

\begin{figure}[htb]
\begin{centering}
\includegraphics[width=0.45\textwidth]{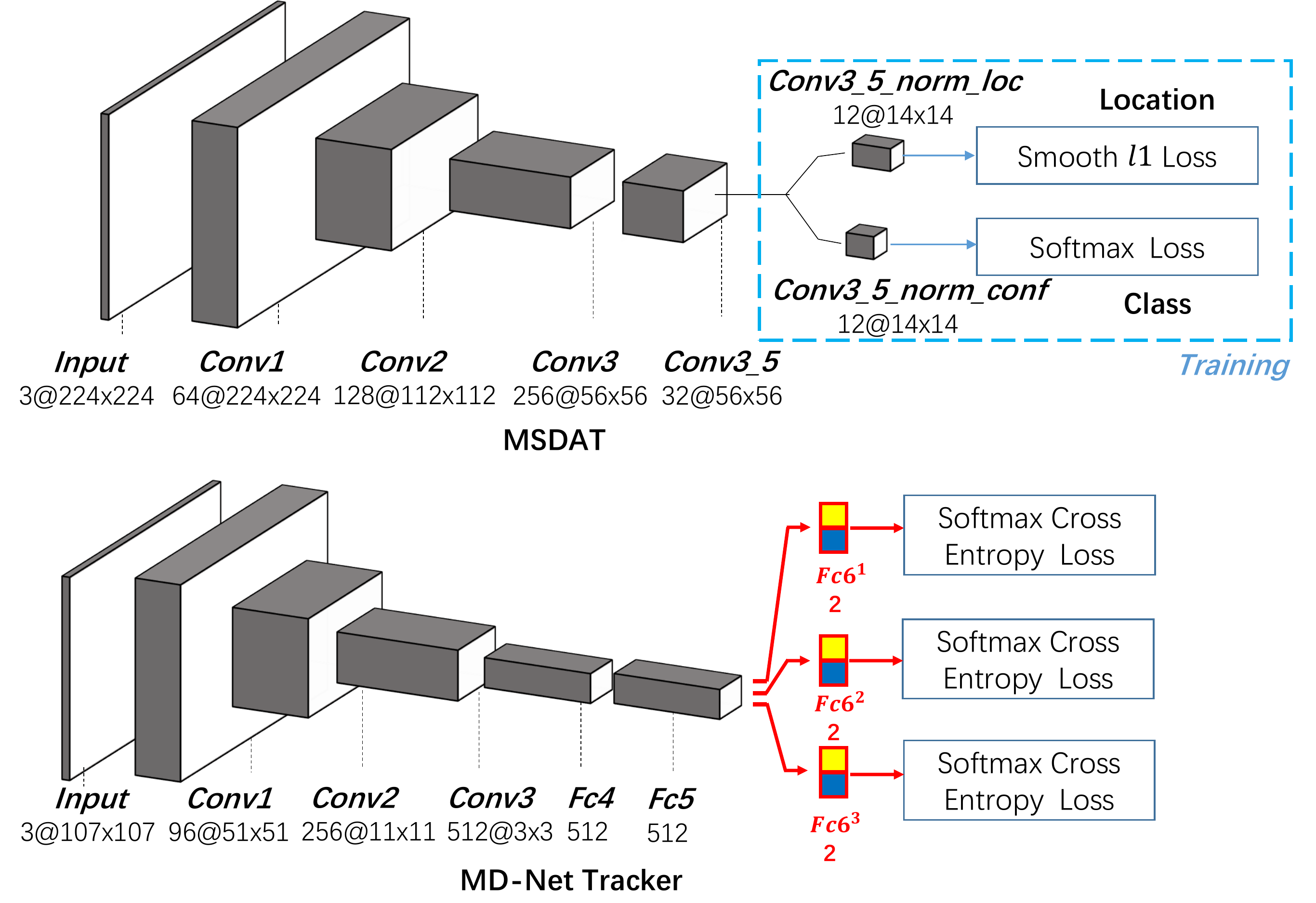}
\par\end{centering}
\caption{
  The flow-charts of the training process of MSDAT and MD-net. Note that the network parts
  inside the dashed blocks are only used for training and will be abandoned before tracking.
  Better viewed in color. 
}  
\label{fig:mdnet_msdat}
\end{figure}

The parameters in the aforementioned tracking branch is learned following a similar manner
as Single Shot MultiBox Detector (SSD), a state-of-the-art detection algorithm
\cite{liu2015ssd}. When training, the original layers of VGG-19 (\emph{i.e.} those ones
before $convx\_5$ are fixed and each ``tracking branch'' is trained independently) The
flowchart of the learning procedure for one tracking branch (based on $conv3\_4$) is
illustrated in upper row of Figure~\ref{fig:mdnet_msdat}, comparing with the learning strategy of
MD-net \cite{nam2015learning} (the bottom row). To obtain a completed training circle, the
adapted feature in $conv3\_5$ is used to regress th objects' locations and their
objectness scores (shown in the dashed block). Please note that the deep learning stage in
this work is purely offline and the additional part in the dashed block will be abandoned
before tracking.

In SSD, a number of ``default boxes'' are generated for regressing the object rectangles.
Furthermore, to accommodate the objects in different scales and shapes, the default boxes
also vary in size and aspect ratios.
Let $m_{i,j} \in \{1, 0\}$ be an indicator for matching the $i$-th default box to the $j$-th
ground truth box. The loss function of SSD writes:
\begin{equation}
  L(m, c, l, g) = \frac{1}{N}\left(L_{conf}(m, c) + \alpha L_{loc}(m, l, g)\right)
\label{equ:tracking_infer}
\end{equation}
where $c$ is the category of the default box, $l$ is the predicted bounding-box while $g$
is the ground-truth of the object box, if applicable. For the $j$-th default box and the
$i$-th ground-truth, the location loss $L^{i,j}_{loc}$ is calculated as 
\begin{equation}
  L^{i,j}_{loc}(l, g) = \sum_{u\in\{x, y, w, h\}}m_{i,j}\cdot\text{smooth}_{L_1}(l^{u}_i -
  \hat{g}^{u}_j)
\label{equ:tracking_infer}
\end{equation}
where $\hat{g}^{u}, u\in\{x, y, w, h\}$ is one of the geometry parameter of normalized
ground-truth box. 

However, the task of visual tracking differs from detection significantly. We thus tailor
the loss function for the KCF algorithm, where both the object size and the KCF window
size are fixed. Recall that, the KCF window plays a similar role as default boxes in SSD
\cite{henriques2015high}, we then only need to generate one type of default boxes and the
location loss $L^{i,j}_{loc}(l, g)$ is simplified as 
\begin{equation}
  L^{i,j}_{loc}(l, g) = \sum_{u\in\{x, y\}}m_{i,j}\cdot\text{smooth}_{L_1}(l^{u}_i -
  g^{u}_j)
\label{equ:tracking_infer}
\end{equation}
In other words, only the displacement $\{x, y\}$ is taken into consideration and there is
no need for ground-truth box normalization.

Note that the concept of domain adaptation in this work is different from that defined in
MD-net \cite{nam2015learning}, where different video sequences are treated as different
domains and thus multiple fully-connected layers are learned to handle them (see
Figure~\ref{fig:mdnet_msdat}). This is mainly because in MD-net samples the training
instances in a sliding-window manner, An object labeled negative in one
domain could be selected as a positive sample in another domain. Given the training video number is $C$ and
the dimension of the last convolution layer is $d_c$, the MD-net learns $C$ independent
$d_c \times 2$ fully-connected alternatively using $C$ soft-max losses, \emph{i.e.}, 
\begin{equation}
  \mathcal{M}^i_{fc}: \mathbb{R}^{d_c} \rightarrow \mathbb{R}^{2}, \forall i = 1 , 2,
  \dots, C
\label{equ:md_mapping}
\end{equation}
where $\mathcal{M}^i_{fc}, \forall i \in \{1, 2, \dots, C\}$ denotes the $C$
fully-connected layers that transferring the common visual domain to the individual object
domain, as shown in Figure~\ref{fig:mdnet_msdat}. 

Differing from the MD-net, the domain in this work refers to a general visual tracking
domain, or more specifically, the KCF domain. It is designed to mimic the KCF input in
visual tracking (see Figure~\ref{fig:mdnet_msdat}). In this domain, different tracking
targets are treated as one category, \emph{i.e.}, objects. When training, the object's location and
confidence (with respect to the objectness) are regressed to minimize the smoothed $l_1$
loss. Mathematically, we learn a single mapping function $\mathcal{M}_{conv}(\cdot)$ as
\begin{equation}
  \mathcal{M}_{msdat}: \mathbb{R}^{d_c} \rightarrow \mathbb{R}^{4}
\label{equ:our_mapping}
\end{equation}
where the $\mathbb{R}^{4}$ space is composed of one $\mathbb{R}^{2}$ space for
displacement $\{x, y\}$ and one label space $\mathbb{R}^2$.

Compared with Equation~\ref{equ:md_mapping}, the training complexity in
Equation~\ref{equ:our_mapping} decreases and the corresponding convergence becomes more
stable. Our experiment proves the validity of the proposed domain adaptation.

\subsection{Multi-scale domain adaptation}
\label{subsec:multi}

\begin{figure}[ht]
\begin{centering}
\includegraphics[width=0.4\textwidth]{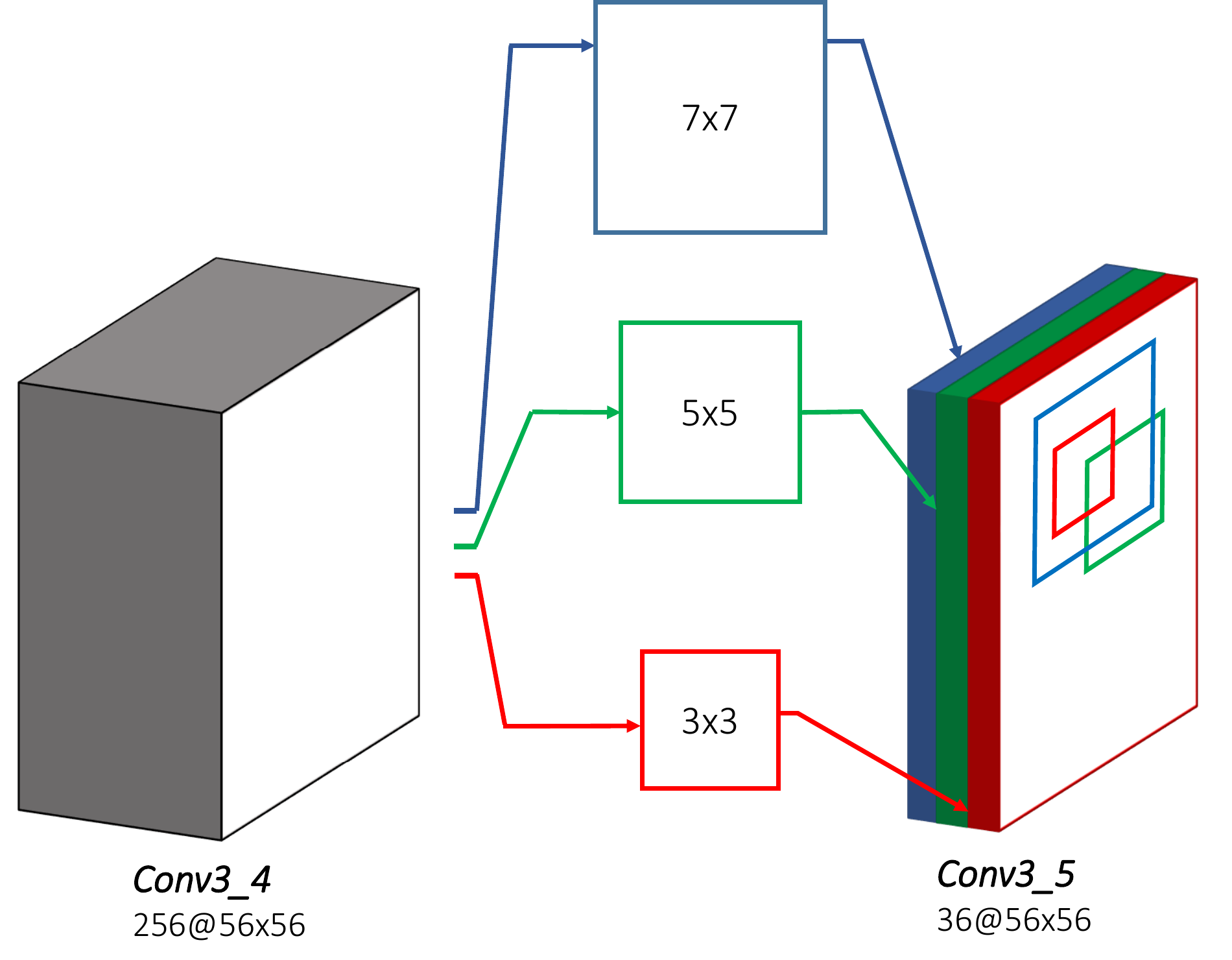}
\par\end{centering}
\caption{
  Learn the adaptation layer using three different types of filters 
}  
\label{fig:inception}
\end{figure}

As introduced above, the domain adaption in our MSDAT method is essentially a convolution
layer. To design the layer, an immediate question is how to select a proper size for the
filters. According to Figure~\ref{fig:structure}, the feature maps from different layers
vary in size significantly. It is hard to find a optimal filer size for all the feature
layers. Inspired by the success of Inception network \cite{szegedy2015going}, we propose to
simultaneously learn the adaptation filters in different scales. The response maps with
different filter sizes are then concatenated accordingly, as shown in
Figure~\ref{fig:inception}. In this way, the input of the KCF tracker involves the deep
features from different scales.

In practice, we use $3\times 3$ and $5\times 5$ filters for all the three feature layers.
Given the original channel number is $K$, each type of filter generate $\frac{K}{16}$
channels and thus the channel reduction ratio is still $8 : 1$.

\subsection{Make the tracker real-time}
\label{subsec:real}

\subsubsection{Channel reduction}
One important advantage of the proposed domain adaptation is the improvement of the
tracking speed. It is easy to see that the speed of KCF tracker drops dramatically as the
channel number increase. In this work, after the adaptation, the channel number is shrunk
by $8$ times which accelerates the tracker by $2$ to $2.5$ times.

\subsubsection{Lazy feed-forward}
\begin{figure}[ht]
\begin{centering}
\includegraphics[width=0.5\textwidth]{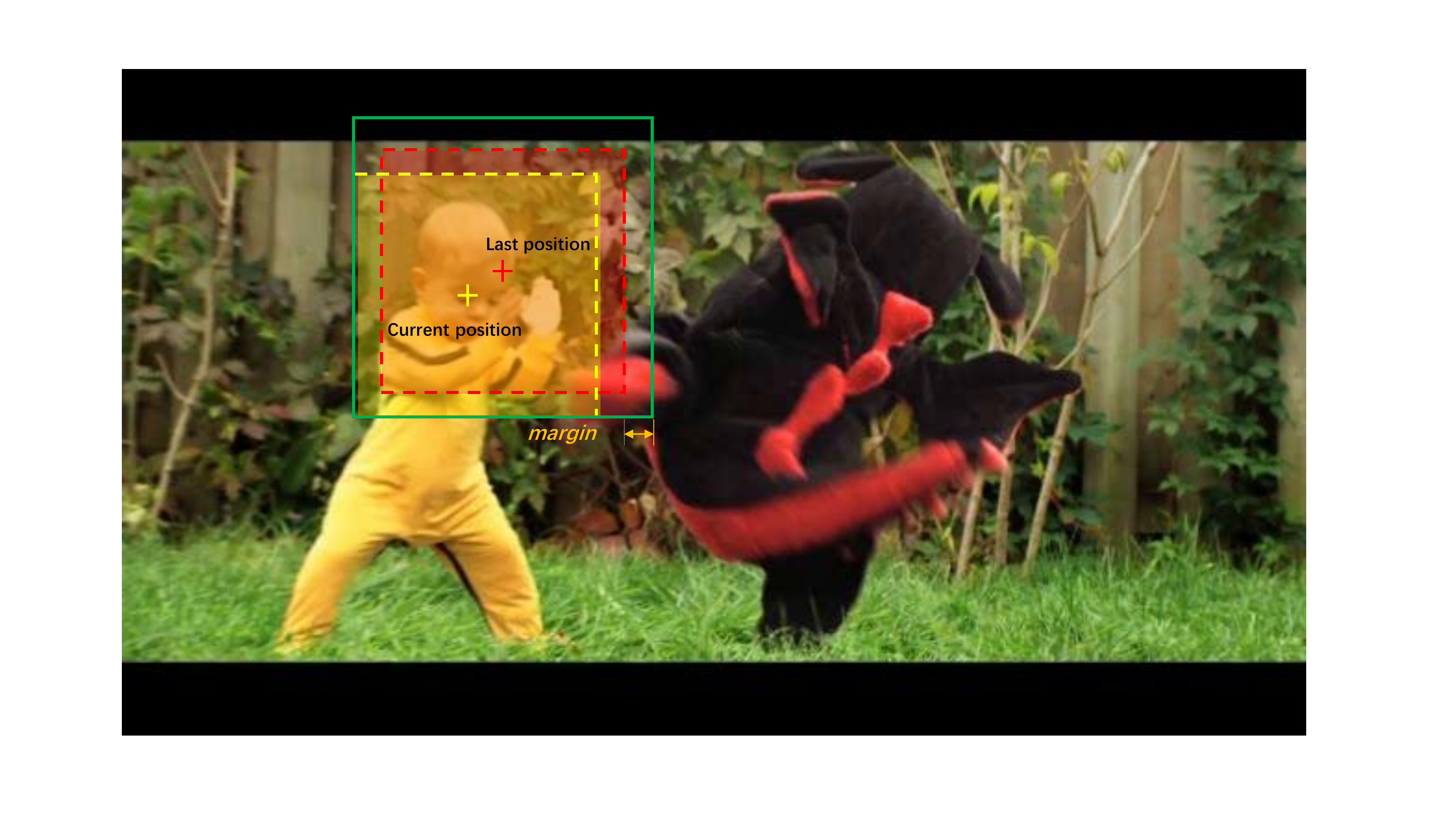}
\par\end{centering}
\caption{
  The illustration of lazy feed-forward strategy. To predict the location of the object
  (the boy's head), a part of the image (green window) is cropped for generating the
  network input. Note that the green window is slightly larger than the red block,
  \emph{i.e.}, the KCF window for predicting the current location. If the predicted
  location (shown in yellow) still resides inside the green lines, one can reuse the deep
  features by cropping the corresponding feature maps accordingly.
}  
\label{fig:lazy}
\end{figure}

Another effective way to increase the tracking speed is to reduce the number of
feed-forwards of the VGG-19 network. In HCF, the feed-forward process is conduct for two
times at each frame, one for prediction and one for model update \cite{ma2015hierarchical}. However, we
notice that the displacement of the moving object is usually small between two frames.
Consequently, if we make the input window slightly larger than the KCF window, one can
reuse the feature maps in the updating stage if the new KCF window (defined by the
predicted location of the object) still resides inside the input window. We thus propose
a lazy feed-forward strategy, which is depicted in Figure~\ref{fig:lazy}. 

In this work, we generate the KCF window using the same rules as HCF tracker
\cite{ma2015hierarchical}, the input window is $10\%$ larger than the KCF window, both in
terms of width and height. Facilitated by the lazy feed-forward strategy, in the proposed
algorithm, feed-forward is conducted only once in more than $60\%$ video frames. This
gives us another $50\%$ speed gain. 

\section{Experiment}
\label{sec:exp}

\subsection{Experiment setting}

In this section, we report the results of a series of experiment involving the proposed
tracker and some state-of-the-art approaches. Our MSDAT method is compared with some
well-performing shallow visual trackers including the KCF tracker
\cite{henriques2015high}, TGPR \cite{gao2014transfer}, Struck \cite{hare2011struck}, MIL
\cite{miltrack}, TLD \cite{kalal2010pn} and SCM \cite{zhong2012robust}. Also, some
recently proposed deep trackers including MD-net \cite{nam2015learning}, HCF
\cite{ma2015hierarchical}, GOTURN \cite{held2016learning} and the Siamese tracker
\cite{bertinetto2016fully} are also compared. All the experiment is implemented in MATLAB
with matcaffe \cite{jia2014caffe} deep learning interface, on a computer equipped with a
Intel i7 4770K CPU, a NVIDIA GTX1070 graphic card and 32G RAM. 

The code of our algorithm is published in Bitbucket via
 \url{https://bitbucket.org/xinke\_wang/msdat}, please refer to the repository for the
implementation details.

\subsection{Results on OTB-50}
\label{subsec:otb50}
\begin{figure*}[thb]
\begin{centering}
\includegraphics[width=0.45\textwidth]{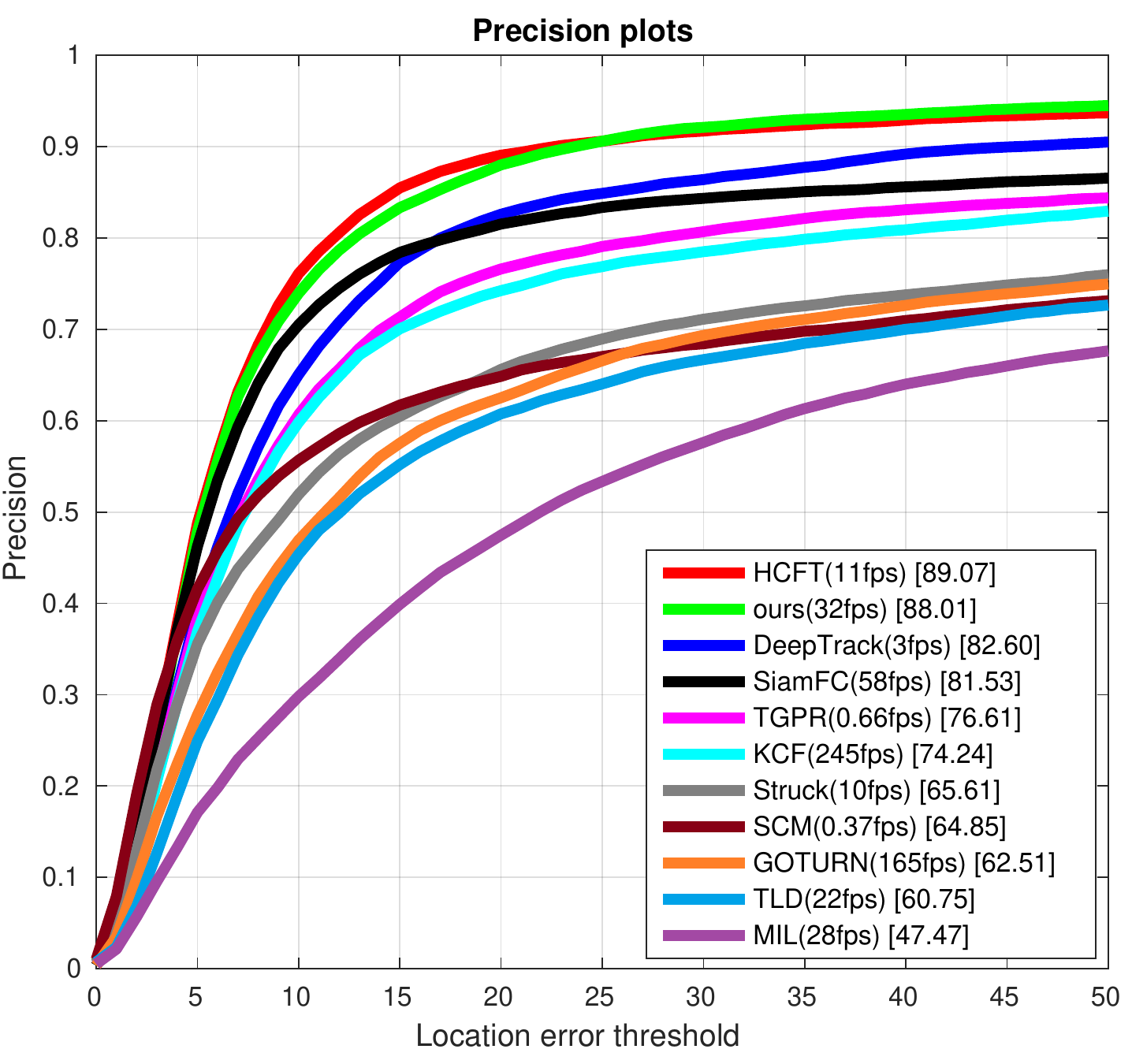}
\includegraphics[width=0.45\textwidth]{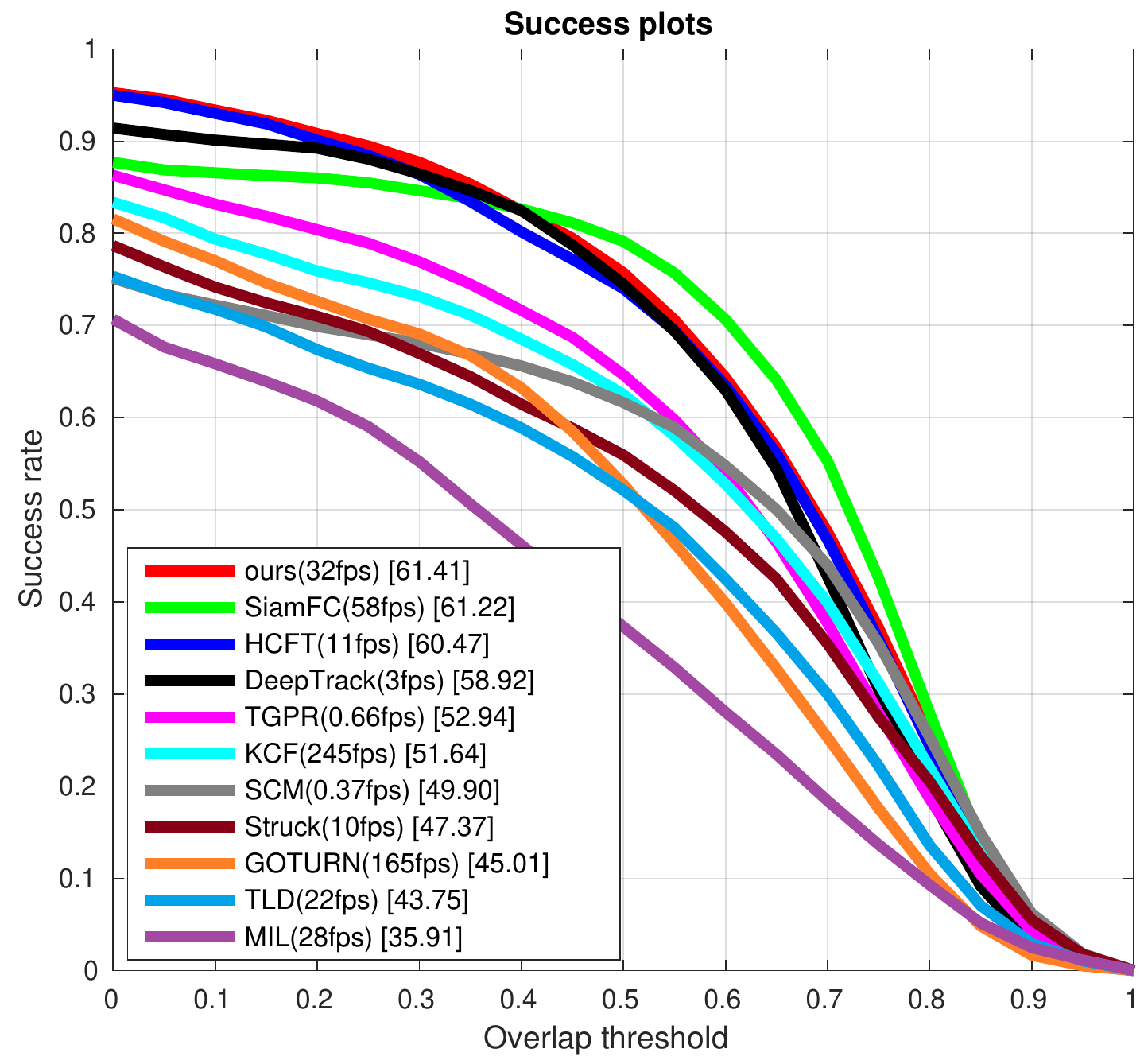}
\par\end{centering}
\caption{
  The location error plots and the overlapping accuracy plots of the involving trackers,
  tested on the OTB-50 dataset.
}  
\label{fig:otb_50}
\end{figure*}

Similar to its prototype \cite{WuLimYang13}, the Object Tracking Benchmark 50 (OTB-50)
\cite{wu2015object} consists $50$ video sequences and involves $51$ tracking tasks.  It is
one of the most popular tracking benchmarks since the year 2013, The evaluation is based
on two metrics: center location error and bounding box overlap ratio. The one-pass
evaluation (OPE) is employed to compare our algorithm with the HCF 
\cite{ma2015hierarchical}, GOTURN \cite{held2016learning}, the Siamese
tracker \cite{bertinetto2016fully} and the afore mentioned shallow trackers. The result
curves are shown in Figure~\ref{fig:otb_50}

From Figure~\ref{fig:otb_50} we can see, the proposed MSDAT method beats all the
competitor in the overlapping evaluation while ranks second in the location error test,
with a trivial inferiority (around $1\%$) to its prototype, the HCF tracker. Recall that
the MSDAT beats the HCF with the similar superiority and runs $3$ times faster than HCF,
one consider the MSDAT as a super variation of the HCF, with much higher speed
and maintains its accuracy. From the perspective of real-time tracking, our method
performs the best in both two evaluations. To our best knowledge, the proposed MSDAT
method is the best-performing real-time tracker in this well-accepted test.

\begin{table*}[tb!]
  \centering
  \resizebox{0.95\textwidth}{!}
  {
	\begin{tabular}{|c|c|c|c|c|c|c|c|c|c|c|c|}
		\hline
		Sequence & Ours & HCF & MD-Net & SiamFC & GOTURN & KCF & Struck & MIL & SCM & TLD\\
		\hline
		DP rate(\%)   & 83.0 & 83.7 & 90.9 &  75.2 & 56.39 & 69.2 & 63.5 & 43.9 & 57.2 & 59.2\\	
		OS(AUC) & 0.567 & 0.562 & 0.678 & 0.561 & 0.424 & 0.475 & 0.459 & 0.331 & 0.445 & 0.424\\
		Speed(FPS) & 34.8 & 11.0 & 1 & 58 & 165 & 243 & 9.84 & 28.0 & 0.37 & 23.3\\
		\hline
	\end{tabular}
  }
  \caption{Tracking accuracies of the compared trackers on OTB-100}	
\label{tab:otb100}
\end{table*}

\subsection{Results on OTB-100}
The Object Tracking Benchmark 100 is the extension of OTB-50 and contains 100 video
sequences. We test our method under the same experiment protocol as OTB-50 and comparing
with all the aforementioned trackers. The test results are reported in
Table~\ref{tab:otb100}

As can be seen in the table, the proposed MSDAT algorithm keep its superiority over all
the other real-time trackers and keep the similar accuracy to HCF. The best-performing
MD-net (according to our best knowledge) enjoys a remarkable performance gap over all the
other trackers while runs in around $1$ fps. 

\subsection{The validity of the domain adaptation}
\begin{figure*}[thb]
\begin{centering}
\includegraphics[width=0.45\textwidth]{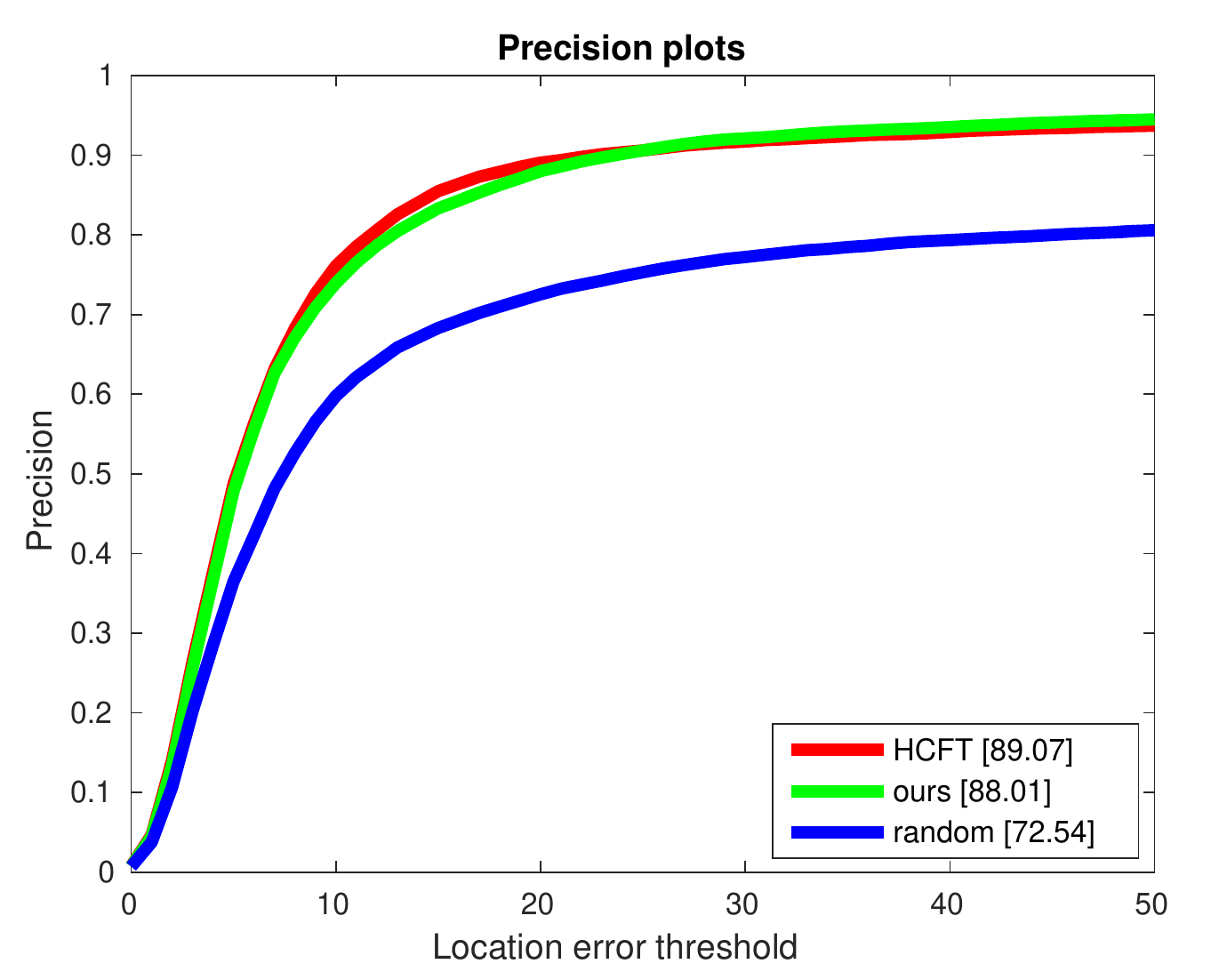}
\includegraphics[width=0.45\textwidth]{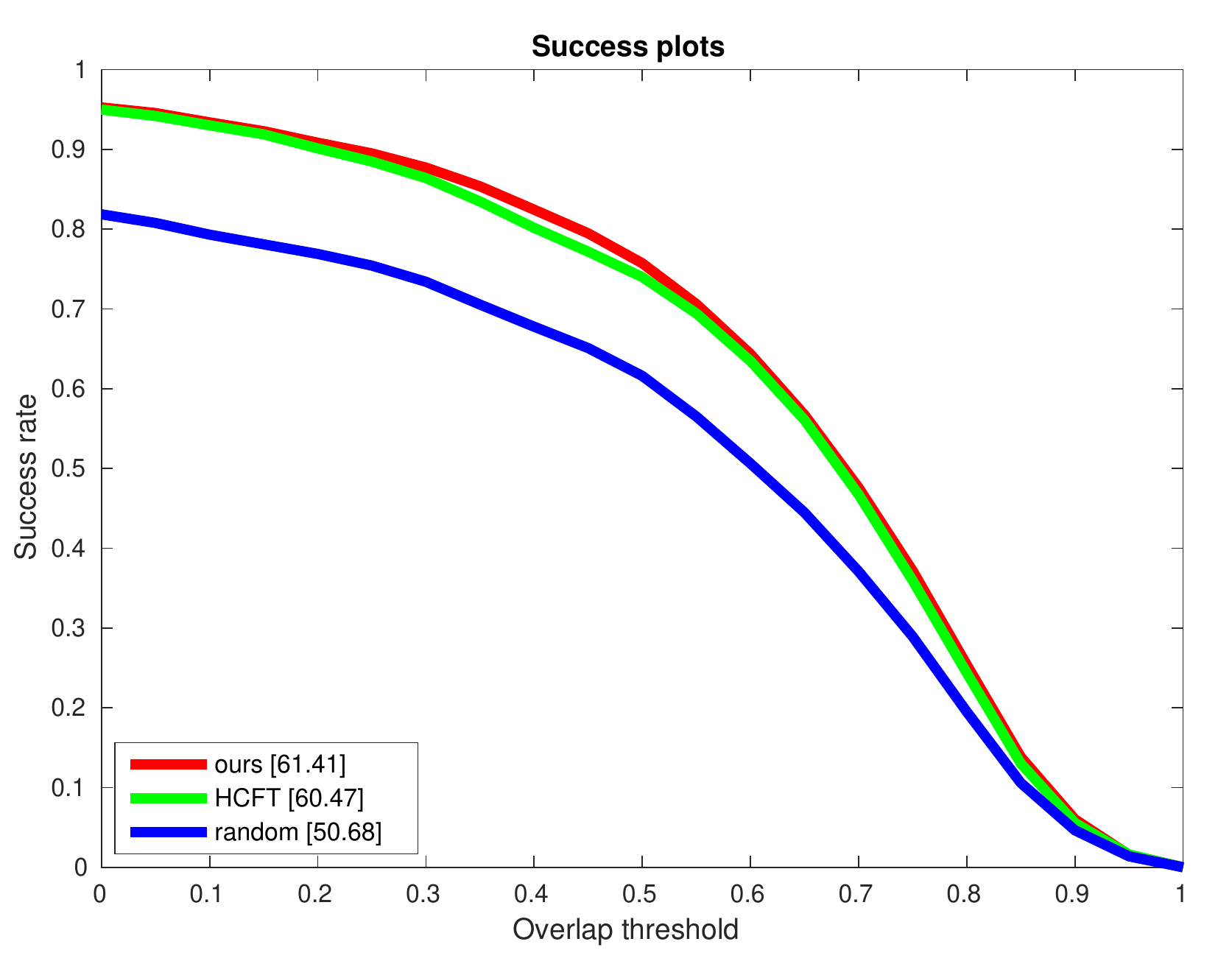}
\par\end{centering}
\caption{
  The location error plots and the overlapping accuracy plots of the three version of the
  HCF tracker: the original HCF, the MSDAT and the random HCF method. Tested on the OTB-50
  dataset, better viewed in color.
}  
\label{fig:rand}
\end{figure*}

To better verify the proposed domain adaptation, here we run another variation of the HCF
tracker. For each feature layer ($conv3\_4$, $conv4\_4$, $conv5\_4$) of VGG-19, one
randomly selects one eighth of the channels from this layer. In this way, the input
channel numbers to KCF are identical to the proposed MSDAT and thus the algorithm
complexity of the ``random HCF'' and our method are nearly the same. The comparison of
MSDAT, HCF and random HCF on OTB-50 is shown in Figure~\ref{fig:rand} 

From the curves one can see a large gap between the randomized HCF and the other two
methods. In other words, the proposed domain adaptation not only reduce the channel
number, but also extract the useful features for the tracking task.

\section{Conclusion and future work}
\label{sec:conc}
In this work, we propose a simple yet effective algorithm to transferring the features in
the classification domain to the visual tracking domain. The yielded visual tracker,
termed MSDAT, is real-time and achieves the comparable tracking accuracies to the
state-of-the-art deep trackers. The experiment verifies the validity of the proposed
domain adaptation. 

Admittedly, updating the neural network online can lift the tracking accuracy
significantly \cite{li2016deeptrack, nam2015learning}. However, the existing online
updating scheme results in dramatical speed reduction. One possible future direction could
be to simultaneously update the KCF model and a certain part of the neural network
(\emph{e.g.} the last convolution layer). In this way, one could strike the balance between
accuracy and efficiency and thus better tracker could be obtained.

\bibliographystyle{IEEEbib}
{\footnotesize
\bibliography{MultiScale_Domain_Adaption.bib}
}

\end{document}